\documentclass{article}
\usepackage{ijcai25}

\usepackage{times}
\usepackage{soul}
\usepackage{url}
\usepackage[hidelinks]{hyperref}
\usepackage[utf8]{inputenc}
\usepackage[small]{caption}
\usepackage{graphicx}
\usepackage{amsmath}
\usepackage{amsthm}
\usepackage{booktabs}
\usepackage{algorithm}
\usepackage{algorithmic}
\usepackage[switch]{lineno}

\usepackage{newtxmath}

\usepackage{tabularray}
\usepackage{multirow}
\usepackage{graphicx}
\usepackage{booktabs}
\usepackage{colortbl}

\title{UniCT Depth: Event-Image Fusion Based Monocular Depth Estimation with Convolution-Compensated ViT Dual SA Block}

\author{
Luoxi Jing$^1$
\and
Dianxi Shi$^{2,1,}$\thanks{Corresponding Author}
\and
Zhe Liu$^2$\and
Songchang Jin$^2$\and 
Chunping Qiu$^2$\and \\
Ziteng Qiao$^2$\and
Yuxian Li$^3$\And
Jianqiang Xia$^4$\\
\affiliations
$^1$School of Computer Science, Peking University\\
$^2$Intelligent Game and Decision Lab (IGDL)\\
$^3$College of Computer, National University of Defense Technology\\
$^4$School of Computer Science, Shanghai Jiao Tong University \\
\emails
jingluoxi@stu.pku.edu.cn,
\{dxshi, liuzhe16, liyuxian\}@nudt.edu.cn,
jsc04@tsinghua.org.cn,
chunping.qiu@aliyun.com,
ztqiao99@163.com,
jianqiang.xia@sjtu.edu.cn
}

\begin{document}

\maketitle

\begin{abstract}
    Depth estimation plays a crucial role in 3D scene understanding and is extensively used in a wide range of vision tasks. Image-based methods struggle in challenging scenarios, while event cameras offer high dynamic range and temporal resolution but face difficulties with sparse data. Combining event and image data provides significant advantages, yet effective integration remains challenging. Existing CNN-based fusion methods struggle with occlusions and depth disparities due to limited receptive fields, while Transformer-based fusion methods often lack deep modality interaction. To address these issues, we propose UniCT Depth, an event-image fusion method that unifies CNNs and Transformers to model local and global features. We propose the Convolution-compensated ViT Dual SA (CcViT-DA) Block, designed for the encoder, which integrates Context Modeling Self-Attention (CMSA) to capture spatial dependencies and Modal Fusion Self-Attention (MFSA) for effective cross-modal fusion. Furthermore, we design the tailored Detail Compensation Convolution (DCC) Block to improve texture details and enhances edge representations. Experiments show that UniCT Depth outperforms existing image, event, and fusion-based monocular depth estimation methods across key metrics.

\end{abstract}

\section{Introduction}

Depth estimation is crucial for understanding 3D scene structures, with broad applications in areas like autonomous driving and medical imaging \shortcite{tang2022perception,wang2024graphcl}.
While image-based methods have achieved significant success, they encounter limitations under extreme lighting conditions, where critical scene details may be lost.
Event cameras, which respond to pixel-level intensity changes, provide advantages such as wide dynamic range and high temporal resolution, making them well-suited for challenging scenarios \shortcite{brandli2014240}. However, the asynchronous and sparse nature of event data poses significant challenges in generating dense predictions from sparse features.
To address these challenges, recent studies have explored event and image modality fusion to combine their strengths \shortcite{gehrig2021combining,shi2023even,pan2024srfnet}. These fusion-based methods utilize the dynamic range and temporal resolution of event cameras alongside the detailed scene information from intensity cameras, achieving more accurate and robust depth estimation in challenging scenarios.

\begin{figure}[t]
    \centering
    \includegraphics[scale=0.5]{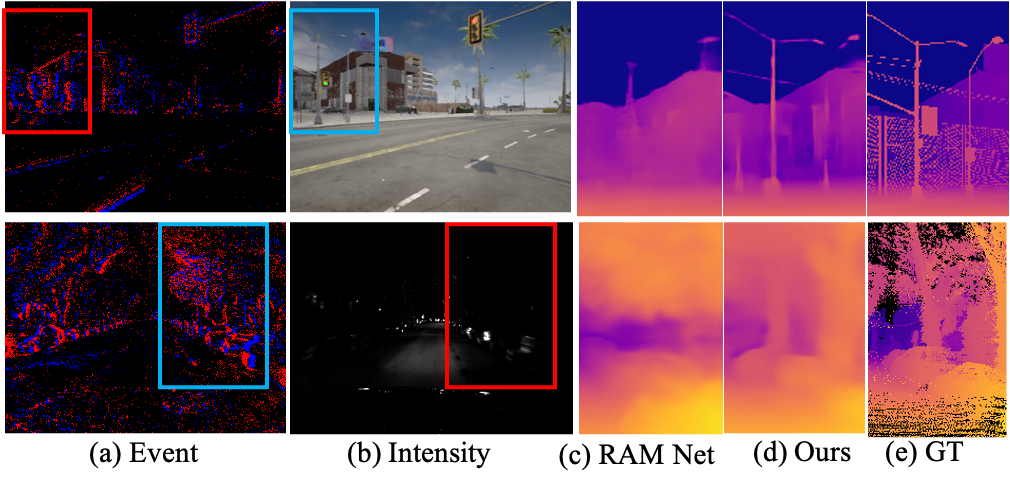}
    \caption{Effects of our methods. Blue boxes highlight objects that exist only in one modality. Our method enhances depth estimation accuracy for occluded areas even in only one modality.}
    \label{fig:intro}
\end{figure}

CNNs are widely used for event-frame fusion in depth estimation tasks \shortcite{gehrig2021combining,shi2023even,zhu2023self}. While effective at capturing local features, they struggle with modeling long-range dependencies, leading to poor performance in scenes with occlusions or significant depth disparities, such as the occlusion between the street lamp and the building shown in Figure \ref{fig:intro}. Some approaches use CNNs with recurrent structures for temporal modeling but face challenges such as gradient vanishing and exploding \shortcite{sutskever2011generating}.
Transformers excel at capturing long-range dependencies. 
Some methods design event representations suitable for Transformer, but they process modality features independently, limiting cross-modal interaction\shortcite{sabater2022event,sabater2023event,hamaguchi2023hierarchical}. SRFNet \shortcite{pan2024srfnet} introduces an attention-based interactive fusion module to merge modality features with spatial priors, but it neglects channel-wise contribution, limiting its effectiveness in fully capturing cross-modal dependencies.
Recently proposed Transformer-based \shortcite{devulapally2024multi} uses a single Transformer encoder to capture cross-modal dependencies. However, it rely on standard self-attention over concatenated modality tokens, resulting in high computational costs and coarse modality fusion, which is susceptible to interference from long-range noise.

In this paper, we propose an event-image fusion-based depth estimation method, UniCT Depth, which adopts a unified conv-transformer architecture to collaboratively model local spatial features and global dependencies. 
In the encoder, we adopt a unified feature extraction and fusion design to reduce redundancy and repetitive computation in traditional separated designs, enhancing cross-modal interaction efficiency and joint representation expressiveness.
We introduce the Convolution-compensated ViT Dual SA (CcViT-DA) Block as the core unit in the encoder. This block integrates Context Modeling Self-Attention (CMSA) branch and Modal Fusion Self-Attention (MFSA) branch to optimize cross-modal collaborative representation. Spatially, CMSA captures contextual dependencies to improve depth estimation in complex scenes, while channel-wise, MFSA captures global dependencies and establishes correlations between modalities, facilitating effective cross-modal fusion. 
To further enhance the CcViT-DA Block, the tailored Detail Compensation Convolution (DCC) block refines local features, improves texture detail extraction, and emphasizes enhanced edge information in multimodal features. The contributions of this work are as follows:

\begin{figure*}[t]
    \centering
    \includegraphics[width=\linewidth,trim=0 310 100 5,clip]{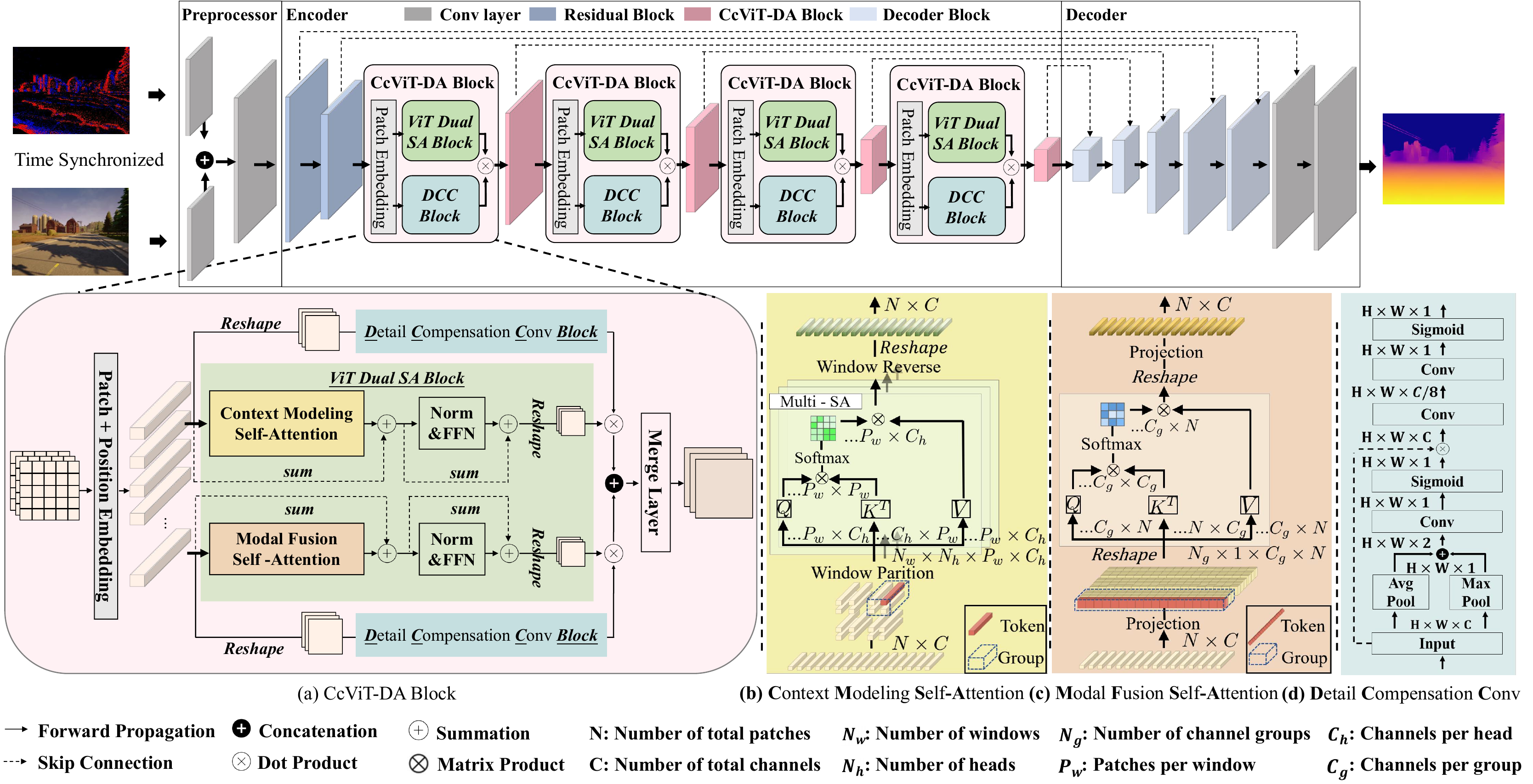}
    \vskip -1.8pt
    \caption{Overview of our proposed UniCT Depth. It processes a time-synchronized pair of event frames and intensity image frames to generate a corresponding depth estimation map. The network architecture comprises three main components. Preprocessor: Extracts and concatenates two modal features from the input data. Encoder: Constructed with residual blocks and CcViT-DA blocks, progressively downsamples the features while extracting high-level semantic representations. Decoder: Upsamples the encoded features and gradually restores spatial resolution, finally producing pixel-wise depth predictions. The skip connection with channel concatenation is used to fuse features between symmetric layers of the encoder and decoder.}
    \label{fig:structure}
\end{figure*}

\begin{itemize}    

    \item We propose UniCT Depth, an event-image fusion depth estimation with a CNN-Transformer architecture that combines local and global feature modeling. Its unified feature extraction and fusion design reduces redundancy and enhances cross-modal interaction for robust depth estimation.

    \item We design CcViT-DA Block to optimize cross-modal representation, integrating CMSA for spatial dependencies and MFSA for modal fusion. The tailored DCC block refines local features, improving depth details and edge representations of concatenated modalities.
    
    \item We conduct experiments using both public real-world datasets and simulated datasets. The results demonstrate that our method outperforms monocular depth estimation algorithms based on events, images, and their fusion, delivering better performance on key metrics.
\end{itemize}

\section{Related Work}

\subsection{Image-based Depth Estimation}

Early image-based methods employed probabilistic and feature-based methods. Yet, these approaches often performed poorly in non-aligned settings, especially when horizontal alignment conditions were not met. 
Learning-based methods, primarily using CNNs, have made significant progress and achieved outstanding results in depth estimation. Eigen et al. \shortcite{eigen2014depth} used a multi-scale convolutional neural network for monocular depth estimation, showing the feasibility of neural networks and inspiring further research into complex models \shortcite{eigen2015predicting,chen2016single}, loss functions \shortcite{jiao2018look,wang2020sdc,ye2019student}, and auxiliary information \shortcite{wang2020cliffnet,lee2020multi} for improved accuracy.
Recently, Transformers have achieved impressive results in computer vision \shortcite{dosovitskiy2020image}.
Ranftl et al. \shortcite{ranftl2021vision} proposed dense prediction transformer (DPT), demonstrating the efficacy of Vision Transformers in dense vision tasks. Since then, transformer-based methods have been widely explored \shortcite{zhao2022monovit,shao2024iebins}.
While image-based methods excel in static or slowly evolving environments, they face considerable challenges under extreme lighting conditions or in the presence of rapidly moving objects.

\subsection{Event-based Depth Estimation}

Model-based methods jointly optimize pose and mapping by solving nonlinear optimization problems, yet typically produce only semi-dense depth \shortcite{kim2016real,rebecq2018emvs,gallego2018unifying}.
Learning-based methods significantly improve the performance of event-based monocular depth estimation, exhibiting strong generalization \shortcite{zhu2019unsupervised,tulyakov2019learning,hidalgo2020learning,shi2023improved}. 
Zhu et al. \shortcite{zhu2019unsupervised} employed a feed-forward neural network to jointly predict camera position and pixel disparity, but it produces only semi-dense depth estimates by applying a mask only to pixels where events occur.
Tulyakov et al. \shortcite{tulyakov2019learning} generated dense metric depth maps by fusing data from stereo setups, while this method still depends on stereo setups and standard feed-forward structures.
Hidalgo-Carrió et al. \shortcite{hidalgo2020learning} proposed a recurrent neural network with temporal consistency supervision, achieving real-time monocular dense depth estimation.
Shi et al. \shortcite{shi2023improved} improved the accuracy of event-based monocular dense depth estimation by utilizing optical flow information between consecutive event frames.
However, event-based methods struggle with high-resolution texture information due to the inherent sparsity of asynchronous event streams and their limited ability to capture scene details.

\subsection{Event-image Fusion Depth Estimation}

Due to the complementary nature of event and image frames, researchers have developed methods to fuse the two modalities. 
CNNs exhibit generalizability in learning local semantics \shortcite{wang2024smooth}, which leads to their widespread adoption for depth estimation in event-image fusion.
Gehrig et al. \shortcite{gehrig2021combining} proposed a fully convolutional recurrent asynchronous multimodal network for depth estimation that can process images and event data. 
Zhu et al. \shortcite{zhu2023self} proposed a self-supervised event-based estimation using cross-modal consistency between aligned frames and events for training.
Shi et al. \shortcite{shi2023even} proposed a three-stage monocular depth estimation framework with a low-light enhancement module, suitable for challenging nighttime conditions. 
CNNs-based methods rely on convolutional networks with limited receptive fields, leading to poor performance in multi-scale or occluded scenes.
Recently, transformer-based methods have gained significant attention in event-image fusion for depth estimation. 
Sabater et al. \shortcite{sabater2022event,sabater2023event} proposed patch-based event representation for transformer architectures.
Hamaguchi et al. \shortcite{hamaguchi2023hierarchical} employed a multi-level memory hierarchy to process event streams, designing an attention-based representation to encode event data.
However, these methods treat each modality independently, leading to suboptimal modality fusion and reduced depth estimation accuracy.
Pan et al. \shortcite{pan2024srfnet} proposed a Spatial Reliability-oriented Fusion, which features an attention-based interactive fusion module that learns consensus regions to guide feature fusion, but it overlooks channel-wise information.
The most relevant work to ours is Transformer-based \shortcite{devulapally2024multi}, which uses a single Transformer to fuse event and image data. However, it employs a basic self-attention mechanism, which is computationally expensive and lacks targeted modality fusion. Furthermore, it relies on ConvLSTM to process events, which is susceptibility to gradient when dealing with long-range dependencies.

In this paper, we propose a CNN-Transformer architecture that combines the strengths of local feature representation and global context modeling, effectively handling multi-scale and occluded complex scenes. In the encoder, we introduce ViT Dual SA block that analyzes the spatial and channel dimensions of modalities. This block not only balances long-range modeling performance and computational efficiency, but also promotes modality fusion by adaptively weighting channels. Besides, we design a DDC Block to enhance the local feature extraction capability, improving the model's ability to capture high-texture objects in complex scenes.

\begin{table*}
    \small
    \tabcolsep=0.1cm
    \centering
    \begin{tabular}{llcccccccccccc|ccc} 
    \toprule
    \multicolumn{2}{c}{\multirow{2}{*}{Method}}                     & \multicolumn{3}{c}{Ourdoor day1}                        & \multicolumn{3}{c}{Ourdoor night1}            & \multicolumn{3}{c}{Ourdoor night2}            & \multicolumn{3}{c|}{Ourdoor night3}           & \multicolumn{3}{c}{Mean Avg.Error}                    \\ 
    \cline{3-17}
    \multicolumn{2}{c}{}                                            & \multicolumn{1}{l}{10m} & 20m           & 30m           & 10m           & 20m           & 30m           & 10m           & 20m           & 30m           & 10m           & 20m           & 30m           & 10m           & 20m           & 30m            \\ 
    \hline
    \multirow{6}{*}{Image based}  & MonoDepth\shortcite{godard2017unsupervised} & 3.44                    & 7.02          & 10.03         & 3.49          & 6.33          & 9.31          & 5.15          & 7.80          & 10.03         & 4.67          & 8.96          & 13.36         & 4.19          & 7.53          & 10.68          \\
                                  & MegaDepth\shortcite{li2018megadepth}        & 2.37                    & 4.06          & 5.38          & 2.54          & 4.15          & 5.60          & 3.92          & 5.78          & 7.05          & 4.15          & 6.00          & 7.24          & 3.25          & 5.00          & 6.32           \\
                                  & MonoVit\shortcite{zhao2022monovit}          & 3.24                    & 5.04          & 5.82          & 3.43          & 5.03          & 6.04          & 3.75          & 4.92          & 5.82          & 4.02          & 4.66          & 5.68          & 3.61          & 4.91          & 5.84           \\
                                  & MonoDEVS\shortcite{gurram2021monocular}     & 1.47                    & 2.49          & 3.13          & 2.99          & 3.71          & 5.08          & 1.77          & 3.17          & 4.66          & 1.40          & 3.01          & 4.68          & 1.91          & 3.10          & 4.39           \\
                                  & DPT\shortcite{ranftl2021vision}             & 1.44                    & 2.40          & 2.82          & 1.80          & 2.67          & 3.22          & 1.68          & 2.59          & 3.06  & 1.57          & 2.45          & 2.94  & 1.62          & 2.53          & 3.01   \\ 
                                  & IEBins\shortcite{shao2024iebins}          & 1.50                    & 2.31          & 2.69          & 2.97          & 3.87          & 4.30          & 2.16          & 3.03          & 3.59          & 1.77          & 2.62          & 3.34          & 2.10          &  2.96         & 3.48           \\
    \hline
    \multirow{4}{*}{Event based}  & Zhu et al.\shortcite{zhu2019unsupervised}   & 2.72                    & 3.84          & 4.40          & 3.13          & 4.02          & 4.89          & 2.19          & 3.15          & 3.92          & 2.86          & 4.46          & 5.05          & 2.73          & 3.87          & 4.57           \\
                                  & DTL-\shortcite{wang2021dual}                & 2.00                    & 2.91          & 3.35          & 2.61          & 3.11          & 3.82          & 1.74          & 2.50          & 3.29          & 1.54          & 2.37          & 3.26          & 1.97          & 2.72          & 3.43           \\
                                  & E2Depth\shortcite{hidalgo2020learning}      & 1.85                    & 2.64          & 3.13          & 3.38          & 3.82          & 4.46          & 1.67          & 2.63          & 3.58          & 1.42          & 2.33          & 3.18          & 2.08          & 2.86          & 3.59           \\
                                  & Mixed-EF2DNet\shortcite{shi2023improved}    & 1.50                    & 2.39          & 2.91          & 2.16          & 2.91          & 3.43          & 1.94          & 2.79          & 3.36          & 1.72          & 2.43          & 2.99             & 1.83             & 2.63             & 3.17              \\ 
    \hline
    \multirow{6}{*}{Fusion based} & RAM Net\shortcite{gehrig2021combining}      & 1.39                    & 2.17          & 2.76          & 2.50          & 3.19          & 3.82          & 1.21          & 2.31          & 3.28          & \underline{1.01}  & 2.34          & 3.43          & 1.53          & 2.50          & 3.32           \\
                                  & EMoDepth\shortcite{zhu2023self}             & 1.40                    & 2.07          & 2.65          & 2.18          & 2.70          & 3.64          & 2.06          & 2.76          & 3.42          & 2.09          & 2.82          & 3.52          & 1.93          & 2.59          & 3.31           \\
                                  & EVT+\shortcite{sabater2023event}  & 1.24        & 1.91          & \underline{2.36}  & 1.45 & 2.10 & 2.88  & 1.48  & \underline{2.13}  & \textbf{2.90}          & 1.38  & 2.03  & \textbf{2.77}        & 1.39  & 2.04  & \underline{2.72}             \\
                                  & HMNet\shortcite{hamaguchi2023hierarchical}  & \underline{1.22}        & 2.21          & 2.68  & 1.50 & 2.48 & 3.19  & 1.36  & 2.25  & 2.96          & 1.27  & 2.17  & 2.86       & 1.34  & 2.28  & 2.92             \\
                                  & Transformer-based\shortcite{devulapally2024multi}  & 1.34             & 2.25          & 2.62  & 1.58 & 2.24 & \textbf{2.78}  & 1.54  & 2.23  & \underline{2.95}          & 1.24  & \underline{1.96}  & 2.81          & 1.43  & 2.17  & 2.79           \\
                                  & SRF Net\shortcite{pan2024srfnet}            & \textbf{0.96}           & \underline{1.77}  & 2.37  & \textbf{1.26} & \textbf{1.95} & 3.01  & \underline{1.19}  & \underline{2.13}  & 3.22          & \underline{1.01}  & 2.12  & 3.52          & \underline{1.11}  & \underline{1.99}  & 3.03           \\
                                  & Ours                                        & \textbf{0.96}           & \textbf{1.74} & \textbf{2.25} & \underline{1.39}  & \underline{1.96}  & \underline{2.86} & \textbf{1.16} & \textbf{1.95} & \underline{2.95} & \textbf{0.84} & \textbf{1.69} & \underline{2.79} & \textbf{1.09} & \textbf{1.84} & \textbf{2.71}  \\
    \bottomrule
    \end{tabular}
    \caption{Quantitative results on the MVSEC dataset. Average absolute depth error (Avg.Error, lower is better) at different cut-off depth distances in meters. The best results are bolded, and the second-best results are underlined. Across all sequences, our method achieves the lowest mean Avg.Error at all cut-off distances, demonstrating both robustness and accuracy.}
    \label{tab:mvsec}
\end{table*}

\section{Method}

In this section, we present our method for estimating dense depth maps from the given image and asynchronous event streams.
We begin by transforming the event stream into an image-like representation. Then we present our network architecture and discuss the loss function used for training the network.

\subsection{Event Representation}

Given the high temporal resolution of event cameras, numerous events can occur in a short span, resulting in a sparse stream. We encode the event stream as a spatio-temporal voxel grid, discretizing the time domain to better preserve temporal information and reduce motion blur \cite{zhu2019unsupervised}. 
Specifically, for an event $e = (x, y,t,p)$, $(x,y)$ denotes the pixel location, $t$ the timestamp, and $p$ the polarity.
The voxel grid is defined as a 3D tensor $\mathbf{V} \in \mathbb{R}^{H \times W \times B}$, where $H$ and $W$ are the height and width of the grid, respectively, and $B$ is the number of time bins. 
For every time bin, the timestamps of occurred events $E_{k} = \{e_i\}_{i=0}^{M-1}$ are scaled to the range $[0,B-1]$. Events are then accumulated to the corresponding voxel grid by bilinear interpolation. The voxel grid $\mathbf{V}_k(x,y,t)$ is defined as follows:

\begin{equation}
    \mathbf{V}_k(x,y,t) = \sum_{i} p_i \delta (x - x_i,y - y_i)\max\{0,1-|t-t_{i}^{*}|\}
\end{equation}

Here, $t_{i}^{*} = \frac{B-1}{\Delta T}(t_i-t_0)$. We set the height and width of the grid to match the resolution of the image.

\subsection{Network Architecture}

As shown in Figure \ref{fig:structure}, our method utilizes a U-Net-like architecture \cite{ronneberger2015u} comprising a preprocessor, encoder, and decoder. 
The preprocessor conducts convolution on both event and image separately, extracting features from each modality to produce full-resolution feature maps. These maps are then concatenated and further convolved to merge features, yielding the full-resolution feature map.

The encoder comprises two residual blocks and four Convolution-compensated ViT Dual Self-Attention (CcViT-DA) modules, enabling efficient communication across multi-scale feature representations. To address the computational load of Transformers with high-resolution images, residual blocks are used to downsample the feature maps, producing half-resolution features. The proposed CcViT-DA modules then serve as the core units to build the encoder.
CcViT-DA block downsamples the feature map by a patch embedding layer, which passes the input feature map through a convolution layer with a $3 \times 3$ kernel and stride of 2, generating a feature map with the resolution halved. In the encoder, the output feature maps have resolutions relative to the input image of $\{\frac{1}{2}, \frac{1}{4}, \frac{1}{8}, \frac{1}{16}, \frac{1}{32}\}$, respectively.

The decoder comprises five decoder blocks. Each decoder block includes a deconvolution layer with a kernel size of 3 and a stride of 2, used to double the size of the input features. In the decoder, the output features have resolutions relative to the input image of $\{\frac{1}{16}, \frac{1}{8}, \frac{1}{4}, \frac{1}{2}, \frac{1}{1}\}$, respectively. The network is designed with skip connections that concatenate features across layers, enhancing the representation of models.

\begin{table*}[tbp]
    \small
    \tabcolsep=0.06cm
    \centering
    \scalebox{0.99}{
    \begin{tabular}{lcccccccccc} 
    \toprule
    \multicolumn{1}{c}{\multirow{2}{*}{Method}} & \multicolumn{5}{c}{Outdoor day1}                                                                                                                                        & \multicolumn{5}{c}{Outdoor night1}                                                                                                                                        \\ 
    \cline{2-11}
    \multicolumn{1}{c}{}                        & Abs. Rel $\downarrow$ & RMSE log $\downarrow$ & $\delta \textless 1.25 $ $\uparrow$ & $\delta \textless 1.25^{2} $ $\uparrow$ & $ \delta \textless 1.25^{3}$ $\uparrow$ & Abs. Rel $\downarrow$ & RMSE log $\downarrow$ & $ \delta \textless 1.25 $ $\uparrow$ & $ \delta \textless 1.25^{2}$ $\uparrow$ & $ \delta \textless 1.25^{3}$ $\uparrow$  \\ 
    \hline
    IEBins\shortcite{shao2024iebins}                          & 0.294                 & 0.497                 & \underline{0.638}                               & \underline{0.839}                                   & \underline{0.932}                                   & 0.503                 & 0.497                 & 0.470                                & 0.697                                   & 0.833                                    \\
    DTL-\shortcite{wang2021dual}                                   & 0.390                 & 0.436                 & 0.510                               & 0.757                                   & 0.876                                   & 0.474                 & 0.555                 & 0.429                                & 0.657                                   & 0.791                                    \\
    E2Depth\shortcite{hidalgo2020learning}                                & 0.346                 & 0.421                 & 0.567                               & 0.772                                   & 0.876                                   & 0.591                 & 0.646                 & 0.408                                & 0.615                                   & 0.754                                    \\
    Mixed-EF2DNet\shortcite{shi2023improved}                              & 0.319                 & 0.389                 & 0.600                               & 0.799                                   & 0.897                                   & 0.428                 & \underline{0.467}         & \underline{0.529}                        & \textbf{0.725}                  & \textbf{0.849}                           \\
    RAM Net\shortcite{gehrig2021combining}                                 & 0.282                 & 0.435                 & 0.548                               & 0.769                                   & 0.871                                   & 0.452                 & 0.537                 & 0.425                                & 0.646                                   & 0.786                                    \\
    Transform-based\shortcite{devulapally2024multi}                 & 0.287         & ——         & 0.351                    & 0.437                           & 0.480                           & 0.348         & ——                 & 0.319                                & 0.440                           & 0.523                                    \\
    SRFNet\shortcite{pan2024srfnet}                                 & \underline{0.234}         & \underline{0.364}         & 0.634                       & 0.814                           & 0.922                           & \underline{0.335}         & 0.544                 & 0.465                                & 0.667                                   & 0.787                                    \\
    Ours                                        & \textbf{0.221}        & \textbf{0.320}        & \textbf{0.665}                      & \textbf{0.853}                          & \textbf{0.934}                          & \textbf{0.311}        & \textbf{0.463}        & \textbf{0.540}                       & \underline{0.722}                           & \underline{0.837}                            \\
    \bottomrule
    \end{tabular}}
    \caption{Detailed comparison of our method with state-of-the-art methods on the MVSEC dataset. $\downarrow$ indicates lower is better and $\uparrow$ higher is better. Our method achieved the best results in 10 out of 12 scores.}
    \label{tab:mvsec_detail}
\end{table*}

\textbf{CCViT-DA Block.} 
The CcViT-DA block comprises a patch embedding layer,  a ViT Dual self-Attention (ViT Dual SA) block, and a Detail Compensation Convolution (DCC) block. Figure \ref{fig:structure} (a)illustrates the architecture of the CcViT-DA block.
We first apply patch embedding and positional embedding to the input feature maps. The patch embedding divides the input into non-overlapping patches, which are linearly transformed into tokens. Positional embedding encodes spatial information. The outputs of both embeddings are then combined to form the input for subsequent processing.

We introduce the ViT Dual SA block with two parallel branches: the Context Modeling Self-Attention (CMSA) branch for contextual dependencies and the Modal Fusion Self-Attention (MFSA) branch for global modal correlations. These branches jointly enhance cross-modal fusion by refining joint representations along spatial and channel dimensions. To further improve local feature representation, we design a DCC Block tailored for the ViT Dual SA block. Specifically, the outputs from both branches are dot-multiplied with the DCC Block outputs, concatenated, and passed through a merge layer to produce the final feature representation.

\noindent\textbf{Context Modeling Self-Attention.} 
The CMSA branch is designed to capture contextual dependencies within the spatial dimension and adapt to local depth variations. By dividing the image into non-overlapping windows, the network focuses on relevant regions within each window, helping to handle depth variations in challenging scenarios like occlusions. 
As shown in Figure \ref{fig:structure}(b), the CMSA block \cite{ding2022davit} calculates multi-head attention within localized windows. The image is partitioned into \( N_w \) non-overlapping windows, each containing \( P_w \) patches, such that the total number of patches \( P \) is \( P = P_w \times N_w \). The attention operation within each window is computed as follows:

\begin{equation}
\begin{split}
    \text{A}_{window}(Q,K,V) &= \{A(Q_i, K_i, V_i)\}_{i=0}^{N_w}, \\
    A(Q_i,K_i,V_i) &= \text{Concat}(head_1, head_2, ..., head_{N_h}), \\
    where \  head_j &= \text{Attention}(Q_i^j, K_i^j, V_i^j)\\
           &= \text{softmax}(\frac{Q_i^j(K_i^j)^T}{\sqrt{C_h}})V_i^j
\end{split}
\end{equation}
where $Q_i, K_i, V_i \in \mathbb{R}^{P_w \times C_h}$ denote the query, key, and value of the $j$-th attention head in the $i$-th window, respectively.
$C_h$ denotes the number of channels for each attention head. 

\noindent\textbf{Modal Fusion Self-Attention.} 
The MFSA branch adaptively models cross-modal feature relationships along the channel dimension, capturing dependencies between modalities. It enhances the contributions of reliable modalities while suppressing noisy features, thereby improving modality fusion and model robustness.
As shown in Figure \ref{fig:structure}(c), MFSA block \cite{ding2022davit} applies the self-attention on a patch-level transposed token, capturing global information along the spatial dimensions by setting the number of attention heads to 1. Each transposed token abstracts the global information. Channels are grouped, and self-attention is applied within these groups to reduce computational complexity. Let $N_g$ denote the number of groups, and $C_g$ denote the number of channels per group, thus $C = N_g \times C_g$. The channel attention mechanism is defined as follows:

\begin{equation}
\begin{split}
        \text{A}_{channel}(Q,K,V) = \{A_{group}(Q_i, K_i, V_i)^{T}\}_{i=0}^{N_g}, \\
        \text{A}_{group}(Q_i,K_i,V_i) = \text{softmax}(\frac{Q_i^T K_i}{\sqrt{C_g}})V_i^T
\end{split}
\end{equation}
where $Q_i, K_i, V_i \in \mathbb{R}^{P \times C_g}$ are grouped channel-wise image-level queries, keys, and values, respectively. 

Unlike the standard self-attention mechanism with quadratic complexity \cite{cui2023strip}, the proposed CMSA block and MFSA blocks reduce the computational complexity from \( O(P^2d) \) to \( O(PP_wd) \) and \( O(PCC_g) \) respectively, significantly improving computational efficiency.

\noindent\textbf{Detail Compensation Conv Block.} We introduce a DCC block that enables the CcViT-DA block to consider both global and local information about the scene. We design a weighted module to facilitate the integration of the convolution and ViT self-attention.
Figure \ref{fig:structure} (d) illustrates the architecture of DCC Block, which aggregates the channel dimensions of the input feature maps to encode spatial regions for emphasis or suppression.
Specially, the channel information of the input features is aggregated using channel-based global maximum pooling and global average pooling, and the two feature maps are concatenated in the channel dimension. Then, they are transformed into a single channel using a convolution layer, and the initial spatial attention map is generated after a sigmoid activation function. The initial spatial attention map are dot product to the input feature map to enhance the input feature space representation. Finally, the feature map is generated as a weighted output with the shape of $H \times W \times 1$, processed through two convolutional layers and activation functions.

\subsection{Loss Function}

The network is trained in a supervised manner, utilizing a loss function that incorporates both L1 and L2 losses. Throughout the training process, the network aims to minimize the loss function at each timestep. Given a sequence of predicted depth maps denoted as $\{D_k\}$, we define the discrepancy $R_k = D_k^* - D_k$, where $D_k^*$ and $D_k$ represent the ground truth and predicted depth values, respectively. The loss function is defined as: $L = \frac{1}{n}\sum_{u}(R_k(x,y) + R_k(x,y)^2)$, where $n$ represents the number of valid pixels in the ground truth depth values.

\begin{figure*}[tb]
    \centering
    \includegraphics[width=1.0\textwidth]{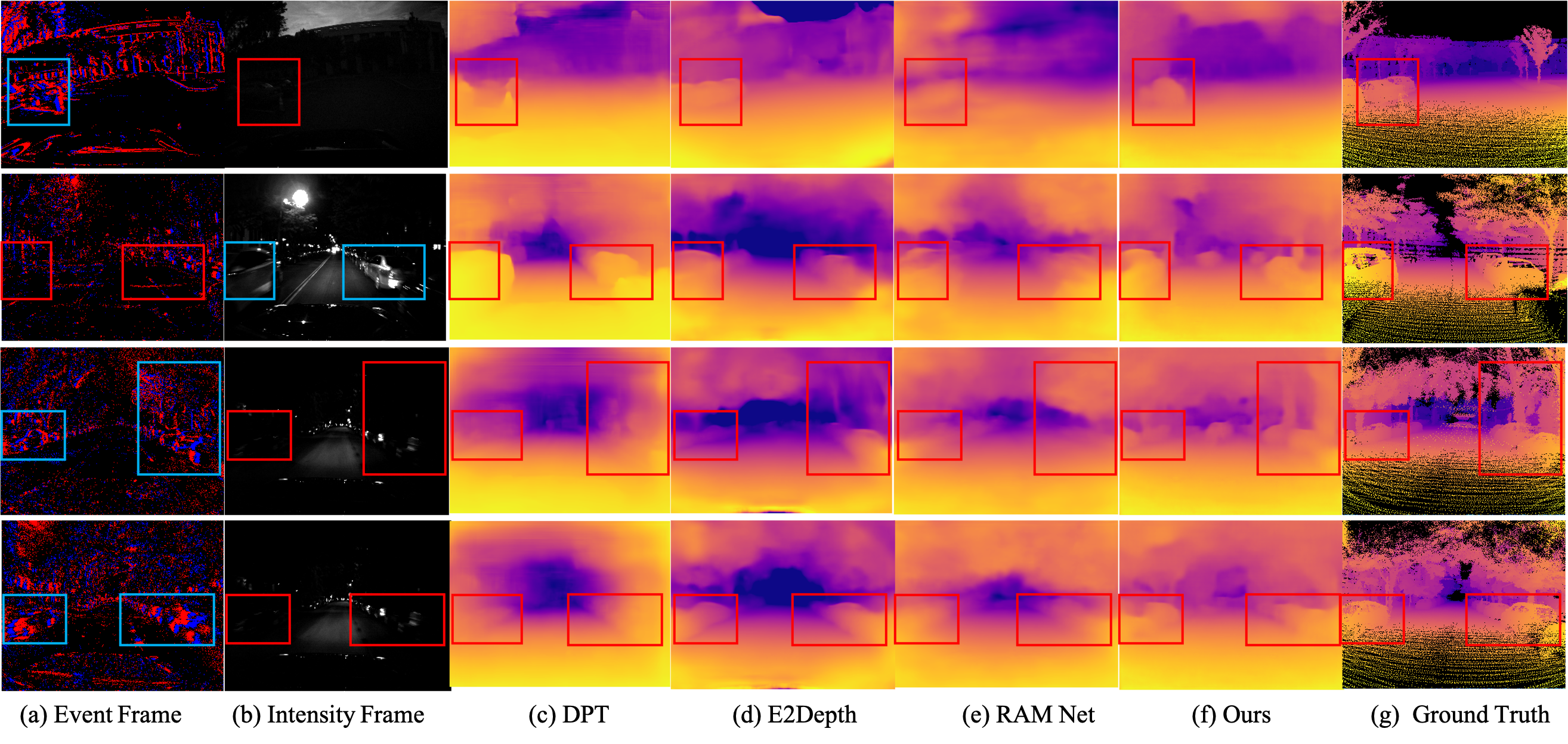}
    \caption{Qualitative comparison for the MVSEC dataset. Compared with baseline methods, our method effectively merges image and event data for more accurate depth estimation and performs well under occlusions.}
    \label{fig:mvsec}
\end{figure*}

\begin{table}[tb]
    \small
    \tabcolsep=0.09cm
    \centering
    \scalebox{0.95}{
    \begin{tabular}{lccccc} 
    \toprule
    \multirow{2}{*}{Method} & \multicolumn{3}{c}{Avg. Error} & \multirow{2}{*}{Abs. Rel $\downarrow$}  & \multirow{2}{*}{RMSE log $\downarrow$}   \\ 
    \cline{2-4}
                            & 10m  & 20m  & 30m              &                           &                            \\ 
    \hline
    DPT\shortcite{ranftl2021vision}                    & 0.53 & 1.04 & \underline{1.75}         & \textbf{0.17}    & 0.34                       \\
    IEBins\shortcite{shao2024iebins}                  & 0.54 & 1.05 & 1.78             & 0.21                      & \underline{0.32}                       \\

    DTL-\shortcite{wang2021dual}                    & 0.84 & 1.46 & 2.16             & 0.26                      & 0.42                       \\
    E2depth\shortcite{hidalgo2020learning}                 & 0.61 & 1.45 & 2.42             & 0.22                      & \underline{0.32}                       \\
    Mixed-EF2DNet\shortcite{shi2023improved}           & \underline{0.30} & 1.23 & 2.18             & 0.19                        & 0.37                         \\
    RAM Net\shortcite{gehrig2021combining}                  & 0.34 & \underline{1.00} & 2.10             & 0.19                      & 0.35                     \\
    Transformer-based\shortcite{devulapally2024multi}                  & 1.04 & 1.87 & 3.65             & 0.22                      & \textbf{0.30}                       \\
    SRFNet\shortcite{pan2024srfnet}                  & 1.50 & 3.56 & 6.11             & 0.51                      & 0.69                       \\
    Ours                    & \textbf{0.26} & \textbf{0.70} & \textbf{1.56} & \underline{0.18}   & 0.36        \\
    \bottomrule
    \end{tabular}}
    \caption{Quantitative results on the DENSE dataset. Our method achieves the best results on all cut-off Avg. Error and performs comparably to the best results in Abs. Rel.}
    \label{tab:dense}
\end{table}

\section{Experiments}

\subsection{Datasets and Evaluation Protocol}
Due to the absence of event data in traditional image datasets like KITTI and NYU, we followed prior work and utilized event camera public datasets for our experiments. Specifically, we conducted primary experiments on the MVSEC dataset \cite{zhu2018multivehicle} to evaluate our method in real-world daytime and challenging nighttime conditions. Besides, we also conducted experiments on the simulated DENSE dataset \cite{hidalgo2020learning} to verify the generalization of methods.

Our model was implemented in PyTorch, utilizing two NVIDIA GeForce RTX 3090 GPUs. 
We choose a learning rate of 0.0002 for MVSEC and 0.002 for DENSE. The ADAMW optimizer and MultiStepLR learning scheduler was used for training, with a batch size of 16. 
The model was trained on the dataset over 50 epochs, with the learning rate being reduced by a factor of 0.5 at the 10th, 20th and 30th epochs. The input image size was configured to 224 $\times$ 224. The number of voxel grid time bins was set to 5, which we found to be a good balance between temporal resolution and computational cost. The weights for the L1 and L2 loss functions were set to 1.

\subsection{Comparison with SOTA Methods}

\textbf{MVSEC dataset.} 
We compare our method with image-based methods, event-based methods, and fusion-based methods on the MVSEC dataset. Following previous work, we evaluate the average absolute depth errors (Avg. Error) of the methods at 10m, 20m, 30m cutoff distances.  
In Table \ref{tab:mvsec}, our method achieves the best performance across all cutoff distances on sequence average.
In Table \ref{tab:mvsec_detail}, we provide a detailed comparison of our method with state-of-the-art approaches on the MVSEC dataset. We employ commonly used metrics in depth estimation, including absolute relative error (Abs. Rel.), logarithmic mean squared error (RMSE log), and accuracy $\delta_n$ ($\delta \textless 1.25^{n}, n=1,2,3$).
Our method achieves the best results in 10 out of 12 scores and remains highly competitive in the remaining two. On the most valuable metric, Abs. Rel., our method improves by 5.56\% and 7.16\% relative to the second-best method, SRFNet, on day and night scenes, respectively. 
Figure \ref{fig:mvsec} presents a qualitative comparison on the MVSEC dataset. In low-light conditions, image-based methods like DPT tend to lose objects, such as trees and cars in the third and fourth rows. Event-based methods lack sufficient texture information, resulting in poor performance in both detail prediction and overall depth estimation of the scene. Compared to the fusion method, RAMNet, our approach effectively merges the two modalities, enabling more accurate depth estimation even for information present in only one modality (e.g., cars in the first row and trees in the third row). Besides, our method outperforms all others in separating foreground and background, especially under occlusions.

\begin{figure*}[tbp]
    \centering
    \includegraphics[width=1.0\textwidth,trim=0 10 0 0, clip]{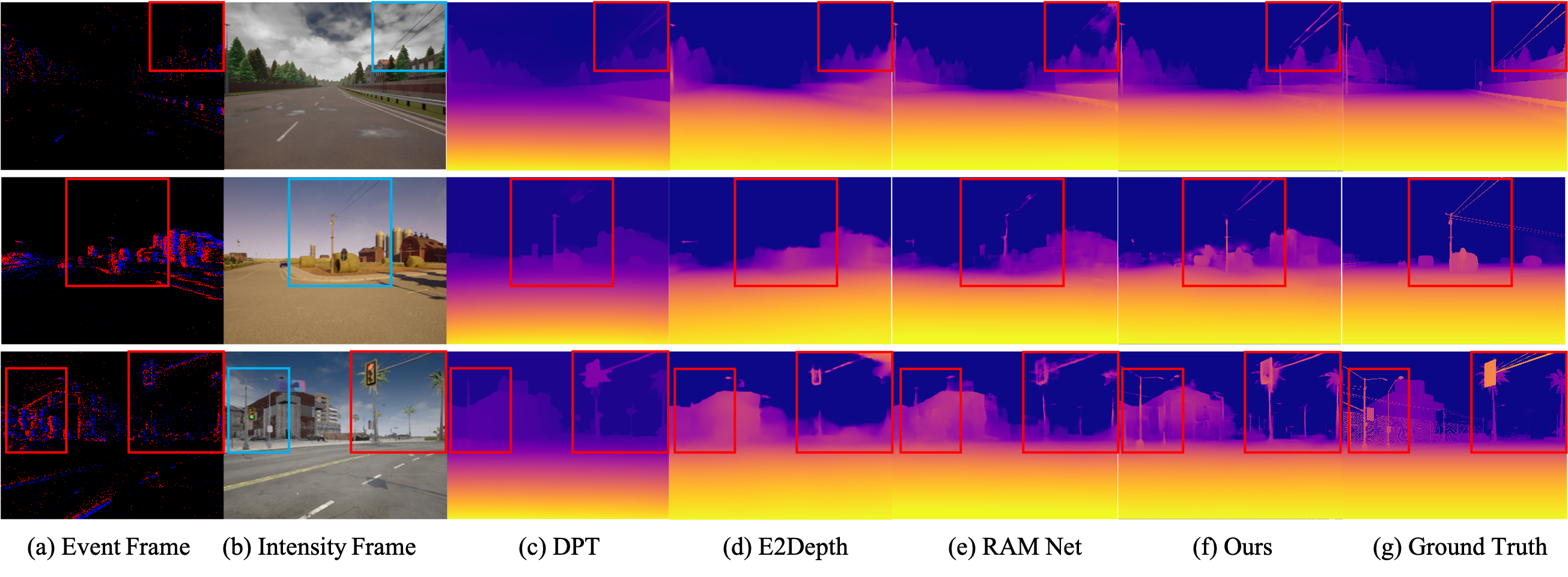}
    \caption{Qualitative comparison for DENSE. Our proposed method provides a more complete and accurate depth estimation of objects, such as trees, telegraph poles, and traffic lights, especially excelling in differentiating foreground objects from the background in occluded regions.}
    \label{fig:dense}
\end{figure*}

\begin{table*}[h]
    \small
    \tabcolsep=0.15cm
    \centering
    \begin{tabular}{ccccccccccc|cccc} 
    \toprule
    \multirow{2}{*}{Method} & \multicolumn{2}{c}{\multirow{2}{*}{ViT Dual SA Block}} & \multicolumn{2}{c}{DCC Block} & \multicolumn{3}{c}{Outdoor day1}              & \multicolumn{3}{c|}{Outdoor night1}           & \multicolumn{3}{c}{Mean Avg.Error}                          & \multirow{2}{*}{FPS $\uparrow$}  \\ 
    \cline{4-14}
                            &   &                & CMSA & MFSA              & 10m           & 20m         & 30m           & 10m           & 20m           & 30m           & 10m            & 20m            & 30m            &                       \\ 
    \hline
    (1)                       & Convolution      & Convolution                                           & $\times$      & $\times$                                           & 1.03          & 2.00          & 2.53          & 1.73          & 2.36          & 3.35          & 1.380          & 2.180          & 2.940          & \textbf{47}                    \\
    (2)                       & Self-Attention      & Self-Attention                                           & $\times$      & $\times$                                           & 1.00          & 1.83          & 2.39          & 1.59          & 2.15          & 3.13          & 1.295          & 1.990          & 2.760          & 13                    \\
    (3)                       & CMSA      & CMSA                                      & $\times$      & $\times$                                           & 1.03          & 2.03          & 2.62          & 1.60          & 2.24          & 3.02          & 1.315          & 2.135          & 2.820          & 22                    \\
    (4)                       & MFSA      & MFSA                                      & $\times$      & $\times$                                           & \underline{0.97}  & 1.80          & 2.41          & 1.47          & 2.16          & 3.02          & 1.220          & 1.980          & 2.715          & \underline{26}                    \\
    (5)                       & MFSA      & CMSA                                      & $\times$      & $\times$                                           & 1.02          & 1.82          & \underline{2.36}  & \underline{1.41}  & 2.05          & 2.94          & \underline{1.215}  & 1.935          & 2.650          & 25                    \\
    \hline
    (6)                       & MFSA & CMSA                                      & $\checkmark$ & $\times$                                           & 1.00          & \underline{1.76}  & \textbf{2.25} & 1.44          & \textbf{1.96} & \underline{2.93}  & 1.220          & \underline{1.860}  & \underline{2.590}  & 25                    \\
    (7)                       & MFSA & CMSA                                      & $\times$      & $\checkmark$                                      & 1.07          & 1.93          & 2.47          & \underline{1.41}  & \underline{1.99}  & 2.99          & 1.240          & 1.960          & 2.730          & 25                    \\
    (8)                       & MFSA & CMSA                                      & $\checkmark$ & $\checkmark$                                      & \textbf{0.96} & \textbf{1.74} & \textbf{2.25} & \textbf{1.39} & \textbf{1.96} & \textbf{2.86} & \textbf{1.175} & \textbf{1.850} & \textbf{2.555} & 25                    \\
    \bottomrule
    \end{tabular}
    \caption{Ablation study on CcViT-DA Block. Evaluate different methods using Avg.Error and FPS on MVSEC. The runtime was measured on RTX 3090 GPU. Our method achieves the lowest error at an acceptable real-time rate.}
    \label{tab:ablation}
\end{table*}

\begin{table}[htbp]
    \small
    \tabcolsep=0.1cm
    \centering
    \scalebox{0.9}{
    \begin{tabular}{ccccccc|ccc} 
    \toprule
    \multirow{2}{*}{Method} & \multicolumn{3}{c}{Outdoor day1}                 & \multicolumn{3}{c|}{Outdoor night1}               & \multicolumn{3}{c}{Average}                       \\ 
    \cline{2-10}
                            & $\delta_1 $ $\uparrow$           & $\delta_2 $ $\uparrow$     & $\delta_3 $ $\uparrow$     & $\delta_1 $  $\uparrow$        & $\delta_2 $ $\uparrow$     & $\delta_3 $ $\uparrow$    & $\delta_1 $ $\uparrow$         & $\delta_2 $ $\uparrow$     & $\delta_3 $ $\uparrow$      \\
    \hline
    E          & 0.602          & 0.816          & 0.918          & \underline{0.483}  & \underline{0.708}  & \textbf{0.853} & 0.542          & \underline{0.762}  & \underline{0.885}   \\
    I      & \textbf{0.684} & \underline{0.845}  & \underline{0.926}  & 0.479          & 0.676          & 0.791          & \underline{0.582}  & 0.761          & 0.859           \\
    Ours                    & \underline{0.665}  & \textbf{0.853} & \textbf{0.934} & \textbf{0.540} & \textbf{0.722} & \underline{0.837}  & \textbf{0.603} & \textbf{0.788} & \textbf{0.886}  \\
    \bottomrule
    \end{tabular}}
    \caption{Ablation study on input modalities. $\delta_n$ denotes the shorthand for $\delta<1.25^n$. E and I represent using only event and intensity image data, respectively. Our method achieves the best results.}
    \label{tab:ablation_modal}
\end{table}

\textbf{DENSE dataset.} 
Table \ref{tab:dense} shows that our method achieves the best results on all cut-off Avg. Error, improving by 13.3\% at 10m, 30\% at 20m, and 10.9\% at 30m compared to the second-best values. Furthermore, our method performs comparably to the best results in Abs. Rel metric.
Figure \ref{fig:dense} shows the comparison of the qualitative results of different methods on the DENSE dataset. 
Compared to the baseline, our method provides a more complete estimation of scene information, demonstrating the effectiveness of model fusion. Additionally, in regions with occlusions, our method more accurately estimates the depth differences between foreground objects and the background. For objects within the scene, such as trees, utility poles, and traffic lights, our method offers more precise depth estimations.

\subsection{Ablation Study}

We conducted an ablation study on the proposed CcViT-DA block. Table \ref{tab:ablation} compares different configurations of the ViT Dual SA and DCC block, with (1) to (5) examining the ViT Dual SA block and (6) to (8) assessing the DCC block. Furthermore, an additional ablation study on the input modalities is presented in Table \ref{tab:ablation_modal}.

\noindent\textbf{ViT Dual Self-Attention Block.} (1), employing a fully convolutional network, achieves the highest frame rate but also the highest error.
(2), which employs traditional self-attention, lowers errors but incurs the lowest frame rate due to significant computation overhead. (3) and (4) improve the frame rate compared to (2). The combination of the MFSA and CMSA branches in (5) results in the lowest errors among configurations (1) to (5) while maintaining an acceptable speed, demonstrating the efficacy of our designed block.

\noindent\textbf{Detail Compensation Convolution Block.} 
(8) outperformed (6) and (7) in terms of Avg. Error across all cut-off depth distances in both day and night scenes, demonstrating the effectiveness of the DCC block. Compared to all other configurations, our method (8) achieved the best depth estimation results while maintaining an acceptable real-time speed. Specifically, compared to (2), the Mean Avg. Error of (8) was reduced by 9.27\%, 7.04\%, and 7.43\% at 10m, 20m, and 30m cut-off distances, respectively. 

\noindent\textbf{Modalities.} Table \ref{tab:ablation_modal} demonstrates that, when used independently, the event modality outperforms the image modality in night scenes, while the image modality achieves superior performance in daytime scenes. However, the combination of both modalities yields the best results in both types of scenes.

\section{Conclusion}

In this paper, we introduce UniCT Depth, a novel monocular depth estimation model that integrates asynchronous event data with images for improved depth accuracy in challenging lighting conditions. Our method combines ViT with CNNs to overcome the shortcomings of traditional CNNs methods, especially in complex multi-scale and occlusion scenarios. It features a dual self-attention block with spatial and channel-wise branches for enhanced pixel relationship modeling and data interaction, significantly enhancing performance. Additionally, a detail compensation convolution boosts local feature extraction, improving the detection of high-texture objects. Our comprehensive experiments on public datasets show that UniCT Depth outperforms existing methods in key metrics. This work improves depth estimation and introduces a new strategy for data fusion across different modalities.

\section*{Acknowledgments}
This work is supported in part by the National Natural Science Foundation of China under Grant 42201513, and in part by the China Postdoctoral Science Foundation under Grant 2022M723902 and Grant 2023T160789.

\appendix

\bibliographystyle{named}
\bibliography{ijcai25}

\end{document}